%% file: acl_latex.tex
\pdfoutput=1

\documentclass[11pt]{article}

\usepackage[final]{acl}

\usepackage{times}
\usepackage{latexsym}

\usepackage[T1]{fontenc}

\usepackage[utf8]{inputenc}

\usepackage{microtype}

\usepackage{inconsolata}

\usepackage{graphicx}

\usepackage[utf8]{inputenc} 
\usepackage[T1]{fontenc}    
\usepackage{hyperref}       
\usepackage{url}            
\usepackage{booktabs}       
\usepackage{amsfonts}       
\usepackage{nicefrac}       
\usepackage{microtype}      
\usepackage{xcolor}         
\usepackage{graphicx}
\usepackage{multirow}
\usepackage{amsmath}
\usepackage{CJKutf8}
\usepackage{algorithm}
\usepackage{algpseudocode}
\usepackage{wrapfig}
\usepackage{pgfplots}
\pgfplotsset{compat=1.17}

\DeclareMathOperator{\PMI}{PMI}
\DeclareMathOperator{\cnt}{count}
\DeclareMathOperator{\prev}{prev}
\DeclareMathOperator{\prefix}{prefix}
\DeclareMathOperator{\PPL}{PPL}

\title{Segment First or Comprehend First?\\Explore the Limit of Unsupervised Word Segmentation with Large Language Models}

\author{
 \textbf{Zihong Zhang\textsuperscript{1}},
 \textbf{Liqi He\textsuperscript{2}},
 \textbf{Zuchao Li\textsuperscript{1,}\thanks{$\ $  Corresponding author. This work was supported by the National Natural Science Foundation of China (No. 62306216), the Natural Science Foundation of Hubei Province of China (No. 2023AFB816) and the Fundamental Research Funds for the Central Universities (No. 2042025kf0026).}},
 \textbf{Lefei Zhang\textsuperscript{2}},
\\
 \textbf{Hai Zhao\textsuperscript{3}},
 \textbf{Bo Du\textsuperscript{2}}
\\
 \textsuperscript{1}School of Artificial Intelligence, Wuhan University, Wuhan, China\\
 \textsuperscript{2}School of Computer Science, Wuhan University, Wuhan, China\\
 \textsuperscript{3}School of Computer Science, Shanghai Jiao Tong University, China
\\
 {\tt \{zhangzihong, heliqi, zcli-charlie, zhanglefei, dubo\}@whu.edu.cn} \\
 {\tt zhaohai@cs.sjtu.edu.cn}
}

\begin{document}
\maketitle
\begin{abstract}
Word segmentation stands as a cornerstone of Natural Language Processing (NLP). Based on the concept of "comprehend first, segment later", we propose a new framework to explore the limit of unsupervised word segmentation with Large Language Models (LLMs) and evaluate the semantic understanding capabilities of LLMs based on word segmentation. We employ current mainstream LLMs to perform word segmentation across multiple languages to assess LLMs' "comprehension". Our findings reveal that LLMs are capable of following simple prompts to segment raw text into words. There is a trend suggesting that models with more parameters tend to perform better on multiple languages. Additionally, we introduce a novel unsupervised method, termed LLACA (\textbf{L}arge \textbf{L}anguage Model-Inspired \textbf{A}ho-\textbf{C}orasick \textbf{A}utomaton). Leveraging the advanced pattern recognition capabilities of Aho-Corasick automata, LLACA innovatively combines these with the deep insights of well-pretrained LLMs. This approach not only enables the construction of a dynamic $n$-gram model that adjusts based on contextual information but also integrates the nuanced understanding of LLMs, offering significant improvements over traditional methods. Our source code is available at \href{https://github.com/hkr04/LLACA}{https://github.com/hkr04/LLACA}
\end{abstract}

\section{Introduction}
\label{intro}

\begin{figure*}[htbp]
\centerline{\includegraphics[width=1\linewidth]{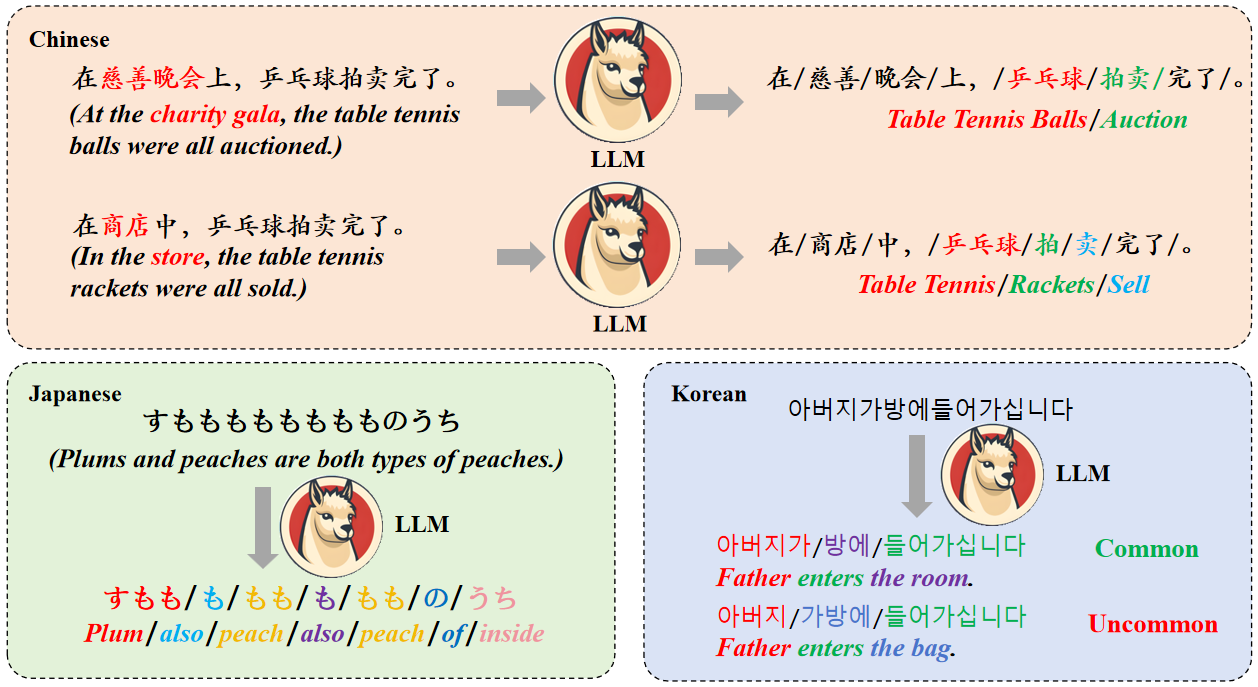}}
\caption{Examples of LLM-WS}
\label{fig:introduction}
\end{figure*}

Understanding language is the core task of NLP~\cite{zhang2019effective}. To measure the understanding capabilities of language models, various Natural Language Understanding (NLU) tasks have been proposed, such as Question Answering~\cite{NLU_QA,yao2023beyond,he2024multi} and Sentiment Classification~\cite{NLU_SC,jing2021seeking}. However, most of these tasks only assess the language model's understanding of the overall meaning of sentences, lacking an evaluation of the language model's understanding capabilities at a fine-grained level. 

In many languages, such as Chinese and Japanese, there are no explicit word boundaries. Therefore, word segmentation is a crucial foundational step in NLP tasks like syntactic analysis~\cite{Syntactic_analysis,li2018unified}, information retrieval~\cite{information_retrieval}, and machine translation~\cite{machine_translation,li2020explicit} for these languages. Most previous research on word segment has adhered to the principle of "segment first, comprehend later", because word segment has long been regarded as the first step in NLP~\cite{zhao@CRFCWS}. However, the human brain's process of analyzing sentences typically involves an interactive process of segmentation and comprehension, where segmentation depends on comprehension and vice versa, especially in sentences with ambiguity. Therefore, word segmentation can also be the last step in NLP, that is, to test the understanding capabilities of a language model.

\textit{Large Language Models (LLMs) are the best linguists.} The emergence of LLMs has marked significant advancements in NLU. And the capabilities of LLMs are no longer based on word segmentation. However, word segmentation can serve as an indicator to effectively evaluate the semantic understanding capabilities of LLMs. Additionally, we can utilize LLMs for word segmentation, and we propose a word segment framework named LLM-Word Segmentation (LLM-WS). We employ the framework to explore the limit of unsupervised word segmentation with LLMs and evaluate the semantic understanding capabilities of LLMs based on word segmentation.

Our research demonstrates that LLMs can perform word segmentation on raw text based on simple prompts, as shown in Figure \ref{fig:introduction}. In the upper section of Figure \ref{fig:introduction}, we demonstrate how LLMs can accurately segment typical Chinese sentences with ambiguity by taking context into account. Based on the different scenarios of "charity gala" and "store", the two identical Chinese word sequences "ping pong balls are auctioned"  and "ping pong racket is sold" were correctly segmented by LLMs. The example in the bottom left corner of Figure \ref{fig:introduction} illustrates that LLMs can correctly segment consecutive identical characters within a sentence. The example in the bottom right corner of Figure \ref{fig:introduction} illustrates that LLMs can segment ambiguous sentences by considering the likelihood of different interpretations within the actual context. Undoubtedly, "father enters the room" is more common than "father enters the bag". The aforementioned examples also demonstrate that LLMs possess capabilities of word segmentation across multiple languages, including Chinese, Japanese, and Korean.

\begin{figure*}[htbp]
\centerline{\includegraphics[width=1\linewidth]{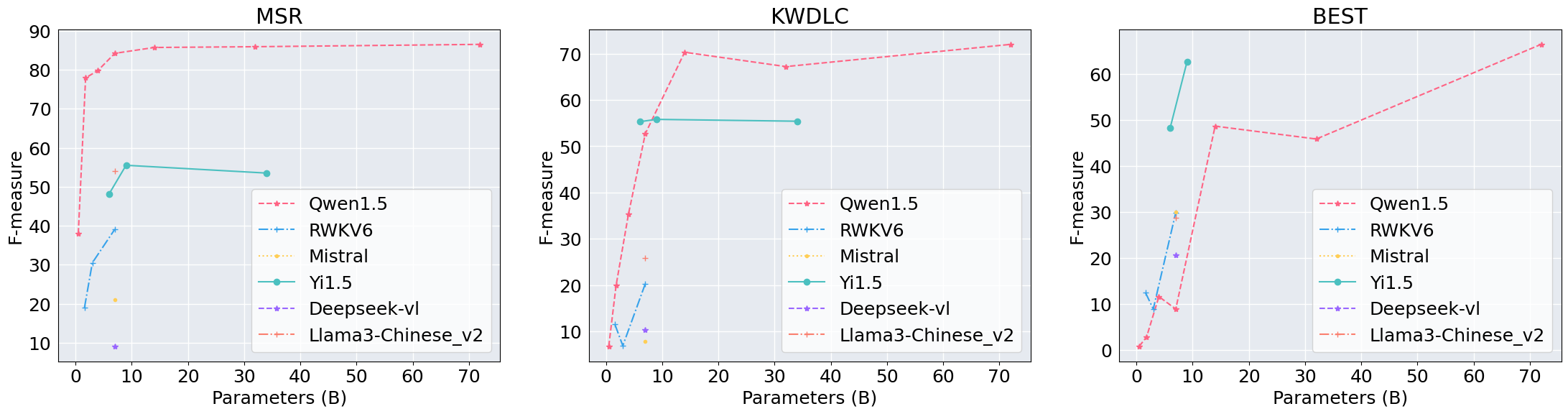}}
\caption{Evaluations on datasets MSR (Simplified Chinese), KWDLC (Japanese) and BEST (Thai).}
\label{fig:pre}
\end{figure*}

In previous word segmentation research, supervised and unsupervised methods are two main learning paradigms for word segmentation. While supervised methods have indeed demonstrated remarkable efficacy~\cite{zhao@CWSreview}, they encounter hurdles such as a heavy reliance on a great deal of manually labeled corpora and poor domain adaptability. Unsupervised word segmentation approaches can avoid the need for a large amount of human labor required for labeled datasets. Additionally, previous research has demonstrated that unsupervised word segmentation methods have better stability for unseen words and adaptability to new domains. Previous unsupervised methods can be broadly classified into two types: discriminative models and generative models. The former evaluates the quality of word candidates using carefully designed goodness measures. However, this method lacks the capability to handle ambiguous strings, which is a major source of segmentation errors. The latter designs models to find the optimal segmentation with the highest generative probability. These methods have better stability in segmenting ambiguous sentences. With the rapid development of neural networks in recent years, research has been conducted to perform unsupervised word segmentation using neural generative models~\cite{wang@SWAN, sun@SLM}, achieving competitive performance to the state-of-the-art statistical models. The most significant difference between our approach and previous research is the implementation of comprehension-based word segmentation. This represents a new era in the development of word segmentation methods. Based on LLMs trained on massive corpora, our framework named LLM-WS can explore the limit of unsupervised word segmentation.

\begin{figure}[htbp]
\centering
\includegraphics[width=0.6\columnwidth]{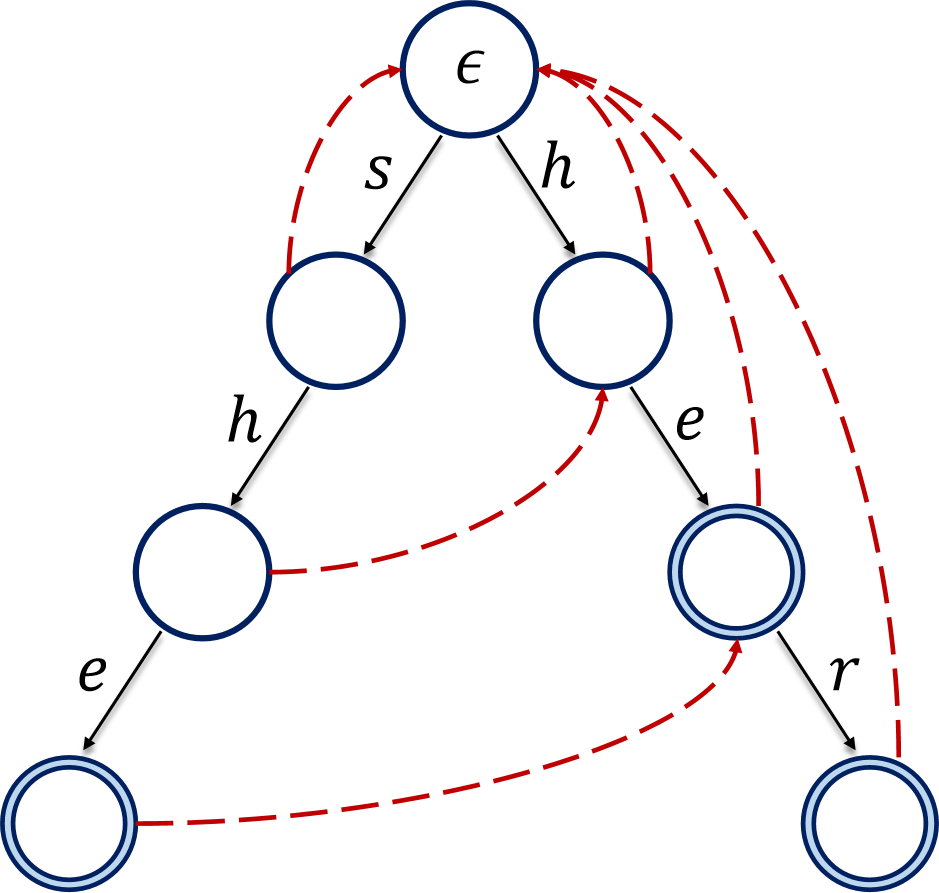}
\caption{The illustration of an Aho-Corasick automaton with patterns "she", "he", and "her", with final states in double circles and failure links in red.}
\label{fig:automaton}
\end{figure}

Specifically, in LLM-WS, we propose an unsupervised word segmentation method named LLACA (\textbf{L}arge \textbf{L}anguage Model-Inspired \textbf{A}ho-\textbf{C}orasick \textbf{A}utomaton). Leveraging the profound insights into language provided by well-pretrained LLMs, we utilize the advanced pattern recognition capabilities of Aho-Corasick automata~\cite{aho1975efficient} to achieve unsupervised word segmentation. We utilize LLMs to conduct word segmentation within the corpus and compute word frequencies. Subsequently, we integrate the computational outcomes derived from LLMs with the AC automaton to facilitate efficient unsupervised word segmentation. 

Our findings reveal that larger language models often exhibit stronger segmentation capabilities. Conversely, models with fewer parameters frequently struggle to adhere to segmentation prompts, leading to significant hallucinations and deviations from human-annotated gold standards. Notably, the Chinese LLM Qwen1.5-7B-Chat has already surpassed previous state-of-the-art results on the Chinese word segmentation tasks for the MSR and PKU datasets.

\section{Related Work}
\label{rel_work}

For word segmentation, the simplest but effective method is the maximum matching model~\cite{jurafsky2014speech}. Beginning at the start of a string, the maximum matching model selects the longest dictionary word that corresponds to the current position and then moves forward to the end of that matched word within the string. However, it is clear that this method cannot recognize words that are not included in the dictionary~\cite{huang@CWSreview}.

With the rise of statistical machine learning methods, word segmentation is formalized as a sequence labeling task. Traditional sequence labeling models such as Hidden Markov Models (HMM)~\cite{carpenter@HMM, yan@HMM-BiMM}, Maximum Entropy Markov Models (MEMM)~\cite{McCallum@MEMM} and Conditional Random Fields (CRF)~\cite{Lafferty@CRF, peng@CRFCWS, zhao@CRFCWS} are widely used. The CRF becomes the mainstream method of supervised word segmentation. Multiple variants of CRF formed the standard word segment models before the deep learning era. 
The linear-chain CRF model is based on the Markov assumption, where the current state depends only on the previous state and the observation sequence, which is not conducive to word segmentation on longer sequences. The first implementation of semi-CRF for word segmentation was published in 2006~\cite{andrew@semiCRF}. Semi-CRF with latent variables was applied to word segmentation, significantly enhancing the performance of CRFs~\cite{sun@semi2009, sun@semi2012}. CRFs can achieve higher word segmentation accuracy, but the complexity of the model leads to slightly lower segmentation efficiency.

\begin{figure*}[htbp]
\centerline{\includegraphics[width=1\linewidth]{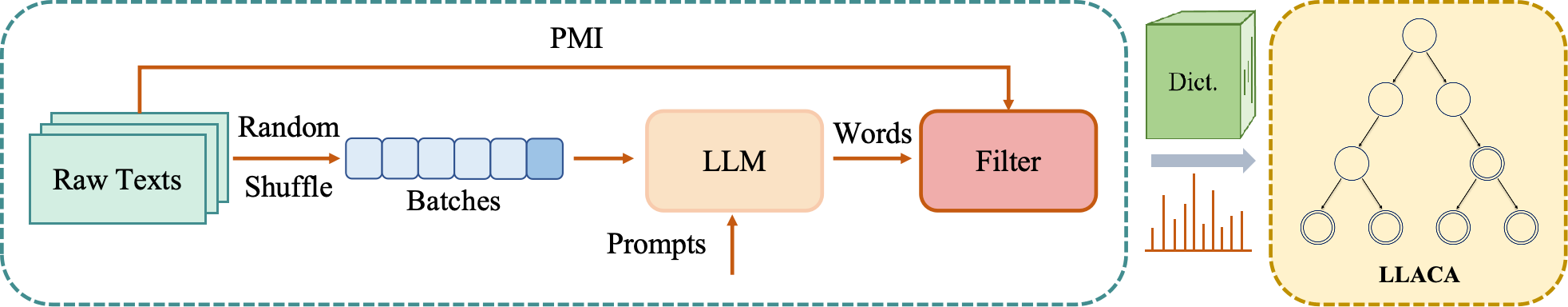}}
\caption{The construction of LLACA.}
\label{fig:LLACA}
\end{figure*}

Most of unsupervised word segmentation methods can be categorized into two types: goodness-based methods and nonparametric Bayesian methods. ESA is a goodness-based method for unsupervised Chinese word segmentation, employing an iterative process with local maximum strategy~\cite{wang@esa}. $n$VBE relies on Variation of Branching Entropy, enhancing performance through normalization and Viterbi decoding while simplifying the model by reducing parameters and thresholds~\cite{magistry@nVBE}. One of the disadvantages of goodness measure based methods is that theoretically, they lack the ability to resolve ambiguity. 
Nonparametric Bayesian methods, such as those proposed by Goldwater et al.~\cite{goldwater@HDP}, a unigram and bigram model based on Dirichlet process and hierarchical Dirichlet process~\cite{teh@hd}. The primary limitation is that the Gibbs sampler requires nearly 20,000 iterations to achieve convergence. Inspired by the "products of experts" idea, a joint model for unsupervised word segmentation combines word-based hierarchical Dirichlet process model with character-based hidden Markov model. The method achieves an unsupervised word segmentation approach that effectively solves ambiguity by calculating the product of probabilities of two generative models and employing Gibbs sampling during the inference process~\cite{chen@joint}.

In recent years, deep neural networks have achieved success in a variety of tasks. Applying methods such as Recurrent Neural Networks (RNNs)~\cite{chen@rnn, sun@SLM} and Long Short-Term Memory Networks (LSTMs)~\cite{cai@lstm, yao@lstm, wang2022unsupervised} to word segmentation can better utilize context and reduce the extensive manual work required for feature engineering. However, neural word segmenters not only require a large amount of training corpora, but also entail more time costs for both training and inference. 

\begin{figure*}[htbp]
\centerline{\includegraphics[width=1\linewidth]{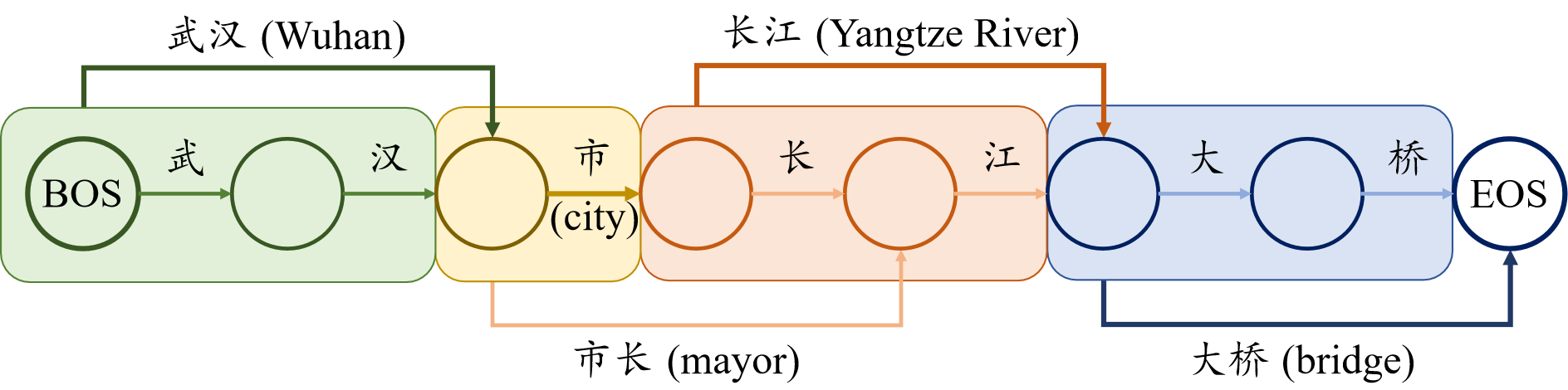}}
\caption{A Directed Acyclic Graph (DAG) for sequence segmentation, starting from "BOS" (Beginning of Sentence) and ending at "EOS" (End of Sentence). Edges denote recognized patterns. Paths through the graph represent potential segmentations, with the preferred path highlighted in frames.}
\label{fig:DAG}
\end{figure*}

\section{Approach}
\label{appr}

To assess the word segmentation capabilities of LLMs, we conducted preliminary experiments using popular open-source LLMs on three datasets from languages without clear word boundaries: MSR~\cite{emerson2005second} for Simplified Chinese, KWDLC~\cite{hangyo-etal-2012-building} for Japanese and BEST~\cite{kosawat2009best} for Thai. Accuracies are quantified using the token F-measure, which is formulated as follows. The F-measure, $F$, is a harmonic mean of precision and recall, where precision ($P$) is the ratio of correctly identified words to the total number of words identified in the output, and recall ($R$) is the ratio of correctly identified words to the total number of words in the gold standard. 

The experiments involved simple prompts directing the LLMs to segment texts into words using spaces without further explanation or contextual information.

\subsection{Preliminary Experiments on LLM-WS}
\label{pre_exp}

The results indicate that models with fewer parameters performed more poorly. Examination of their outputs revealed two primary issues: firstly, these models completely failed to comprehend the prompts, resulting in chaotic outputs that included repeated or incorrectly replaced words; secondly, although some models partially understood the prompts and attempted text segmentation, various misalignments occurred, such as inappropriate translations and misuse of delimiters.

LLMs with more parameters, like 7 billion, generally demonstrated an adequate understanding of the task when processing texts in languages consistent with their training corpus. As the parameter count increased, hallucinations diminished, and F scores improved across different languages.

\subsection{LLACA: Large Language Model-Inspired Aho-Corasick Automaton} 

The Aho-Corasick automaton (AC automaton), as defined by \citet{aho1975efficient}, extends the concept of a Trie, a tree-like data structure commonly used for storing strings where each node signifies a string prefix, and each directed edge represents an input character or byte. Patterns are incorporated into the Trie, with each path from the root to a node denoting a pattern prefix. Notably, if a node's path represents an entire pattern, it is designated as a final state. Each node in the Trie has added failure links directing the automaton to the next node to consider if a character does not match at the current node, effectively allowing the automaton to skip unnecessary nodes and thus speeding up the pattern-matching process. Figure \ref{fig:automaton} illustrates a typical AC automaton configured with patterns "she", "he", and "her". Both the construction and recognition processes of the automaton operate with linear time complexity relative to the length of the input text. 

Traditionally, methods like maximum length matching and maximum probability matching for word segmentation have leveraged AC automata in conjunction with human-annotated dictionaries. However, the dynamic nature of language and the considerable effort required for human annotations present substantial challenges.

To address these issues, our approach integrates the deep comprehension capabilities of LLMs to develop vocabularies that requires no human annotation. By harnessing the understanding inherent in LLMs, we can dynamically adapt to language changes over time, significantly reducing the labor and limitations associated with manual methods. This fusion not only enhances the automaton's application scope but also paves the way for more autonomous and robust text processing tools.

The initial step involves sampling from a LLM. We begin by randomly shuffling the raw sentences into several batches, each approximately the size of the square root of the total number of sentences. For each batch, we use simple prompts to direct the LLM to perform LLM-WS. This process may occasionally lead to errors, such as the LLM mistaking the task for translating into its primary training language, commonly referred to as "hallucinations". To address this, a filtering step is crucial. We employ Pointwise Mutual Information (PMI) to assess the coherence of each "word" $w[1..n]$ identified by the LLM. In Equation \ref{eqn:pmi}, $p(\cdot)$ is computed as a unigram probability. 

\begin{equation}
\small
    \PMI(w) = \min_{i=1}^{n-1}\left(\log\frac{p(w[1..n])}{p(w[1..i])\cdot p(w[i+1..n])}\right)
    \label{eqn:pmi}
\end{equation}

\input{tab/main.tex}

An important hyperparameter in this context is the "top ratio", which determines the proportion of words to retain. To minimize the risk of missing out-of-vocabulary (OOV) items, this parameter should not be set too low. 

Figure \ref{fig:LLACA} illustrates the construction of LLACA. Once constructed, it analyzes the semantic patterns offered by the LLM-WS. Unlike the widely-used open-sourced Chinese word segmentation system Jieba\footnote{\url{https://github.com/fxsjy/jieba}}, which models the probability of a word occurring as $p(w)=\frac{\cnt(w)}{\sum \cnt(w_i)}$, we adopt a different approach. We define the previous state $\prev(w)$ as the closest final state on the path from the root to the current state $w$. For example, as shown in Figure \ref{fig:automaton}, the previous state of "her" is "he". And the root, representing $\epsilon$, is considered a special final state. Thus the previous state of "she" is $\epsilon$. The denominator is now defined as the sum of word counts in the sub-Trie of $\prev(w)$. In other words, the probability of $w$ is affected by its occurrence count and those of others with the same prefix $\prev(w)$. 

\begin{equation}
    p(w) = \frac{\cnt(w)}{\sum\limits_{\prev(w)\in \prefix(w_i)} \cnt(w_i)}
    \label{eqn:trans_prob}
\end{equation}

When no other pattern serves as a prefix for a given pattern, the model reverts to using unigram probabilities. When the longest prefix of a word 
$w$, such as "he" for "her", reaches a final state, the model effectively functions as an $n$-gram model, where $n$ represents the length of the word $w$. This approach dynamically captures contextual information at the character level using the variable $n$-gram model and extends this context into Viterbi decoding at the word level.

For example, suppose we have patterns "a", "ab", "ac", "bc" with counts $1,2,3,4$, then we can calculate the probility of how possible "ab" could be a word as $\frac{2}{1+2+3}$ since it's longest recognizable prefix is "a", which is the common prefix of "a", "ab" and "ac". And for "bc", since there's no "b" in the patterns, it's calculated as $\frac{4}{1+2+3+4}$ because all of the patterns share common prefix $\epsilon$ denoting the empty string.

Ambiguities within words, such as "\begin{CJK}{UTF8}{gbsn}武汉市长\end{CJK}" (the mayor of Wuhan city), might be wrongly segmented into "\begin{CJK}{UTF8}{gbsn}武汉市/长\end{CJK}" (Wuhan city/long) by simple unigram models like Jieba, where both "\begin{CJK}{UTF8}{gbsn}武汉市\end{CJK}" (Wuhan city) and "\begin{CJK}{UTF8}{gbsn}长\end{CJK}" (long) are frequent patterns in Chinese. Our model addresses these situations more adeptly by leveraging contextual cues: "\begin{CJK}{UTF8}{gbsn}武汉\end{CJK}" yields a lower perplexity in the context following "\begin{CJK}{UTF8}{gbsn}武\end{CJK}", and similarly, "\begin{CJK}{UTF8}{gbsn}市长\end{CJK}" (mayor) is more likely following "\begin{CJK}{UTF8}{gbsn}市\end{CJK}" (city). Thus, in Viterbi decoding, our model can more accurately segment it as "\begin{CJK}{UTF8}{gbsn}武汉/市长\end{CJK}" (the mayor of Wuhan city) rather than "\begin{CJK}{UTF8}{gbsn}武汉市/长\end{CJK}" (Wuhan city is long).

We adopt a 2-tag system, where each decision either marks a boundary between words or allows the sequence to continue without a break. In Viterbi decoding, we traverse the sequence from beginning to end, updating the path with the maximum log probability for every prefix. Each position stores the optimal previous word boundary and its corresponding log probability. Once the path for the last prefix (the whole sequence) is updated, we can backtrack from it to the beginning of the sentence to determine all the word boundaries in the optimal path.

\input{tab/qwen_gpt.tex}

\subsection{Time Complexity} 

The LLM-WS will be the most time-consuming part, primarily due to the inference efficiency of the LLM. Apart from that, every component of LLACA is robust and quick. PMI can be calculated in $\mathcal{O}(L^2\cdot \log N)$, where $L$ denotes the length of the word being analyzed, and $N$ denotes the length of the raw text using a Suffix Array~\cite{manber1993suffix} built on the raw text. The construction of the AC automaton and the pattern matching are linear with respect to the sum of the lengths of the patterns and the raw text~\cite{aho1975efficient}, respectively. The time complexity of Viterbi decoding~\cite{viterbi1967error}, which involves traversing the Directed Acyclic Graph (DAG) that represents all recognized patterns (as shown in Figure \ref{fig:DAG}), is approximately $\mathcal{O}(N)$, depending on the number of patterns matched. More details are available in Appendix \ref{exp_dtl}.

\section{Experiments}
\label{exp}

\subsection{Experimental Setup} To evaluate the word segmentation capabilities of LLMs and the effectiveness of our proposed approach across different languages and scripts, we selected several open-sourced datasets for Chinese (AS, CITYU, MSR, PKU,~\citealp{emerson2005second}), Japanese (KWDLC,~\citealp{hangyo-etal-2012-building}, UD\_JA,~\citealp{nivre2020universal}), Korean (UD\_KO,~\citealp{nivre2020universal}) and Thai (BEST,~\citealp{kosawat2009best}, UD\_TH,~\citealp{nivre2020universal}). We continued to use the token F-measure as our primary metric for evaluation, consistent with the methodology outlined in our preliminary experiments (\ref{pre_exp}). 

The sampling and testing procedures were conducted on the test sets. Except for the BEST dataset, which was randomly sampled from the training data, all other test datasets maintained their original splits. Our experiments primarily utilized the Qwen1.5 series of LLMs~\cite{bai2023qwen}. We selected this series because it is reputed to excel in multilingual tasks, offering a wide range of parameters from 0.5B to 110B and easy to employ. The diversity of parameters enables us to explore the relationship between the models' parameter sizes and their comprehension abilities effectively. For further details about our experimental setup, please see Appendix \ref{exp_dtl}.

\begin{figure*}[htbp]
\centerline{\includegraphics[width=1\linewidth]{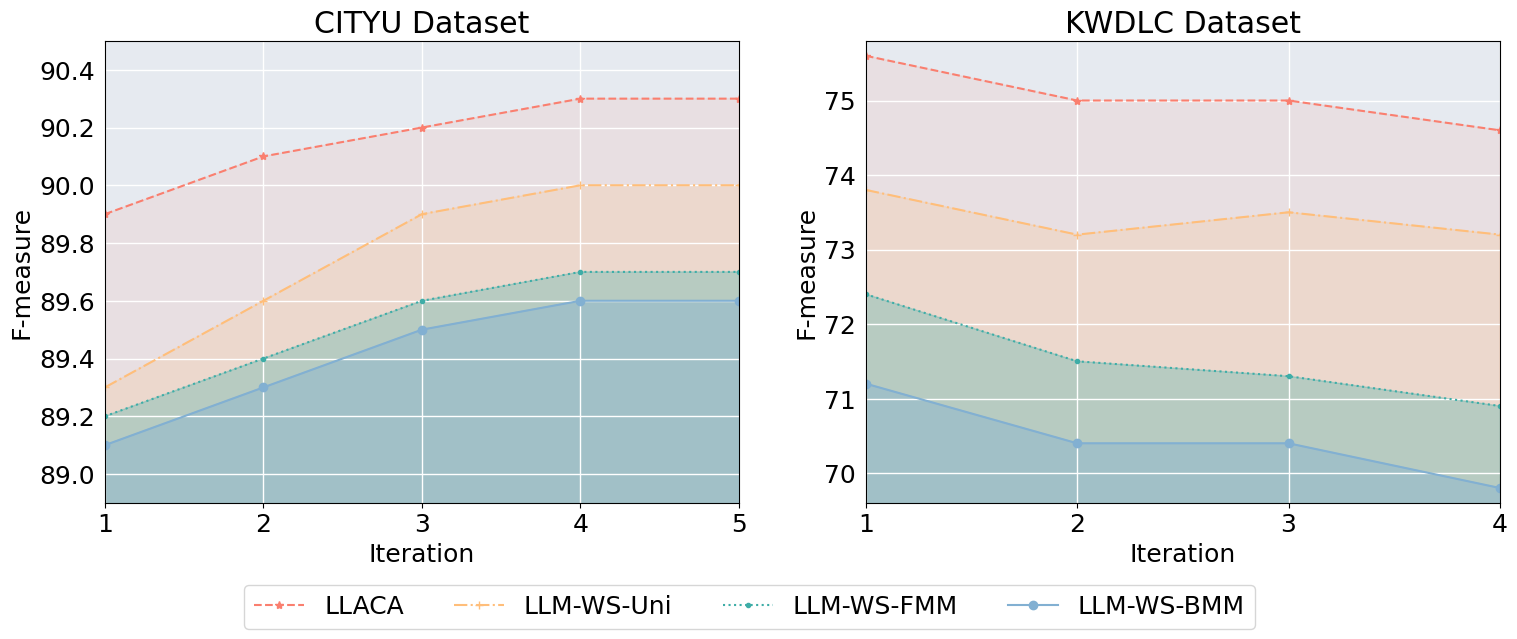}}
\caption{The evolution across iterations for four methods based on LLM-WS, starting from a baseline of 86.8 for the CITYU dataset and 72.3 for the KWDLC dataset with the Qwen1.5-110B-Chat model.}
\label{fig:comp}
\end{figure*}

\subsection{LLM-WS: Evaluating the Word Segmentation Capabilities of LLMs} 
Table \ref{tab:res} presents the results after one iteration and compares them with previous state-of-the-art unsupervised methods. Previous state-of-the-art unsupervised methods for Chinese word segmentation generally achieved scores around 80. However, Qwen1.5-7B-Chat, which was mainly pre-trained on Chinese corpus, has significantly advanced performance, reaching approximately 87. It is important to note that Qwen was primarily pre-trained on Simplified Chinese; consequently, it occasionally produces Simplified Chinese outputs when processing Traditional Chinese. To address this, texts are converted to Simplified Chinese using OpenCC\footnote{\url{https://github.com/yichen0831/opencc-python}}, resulting in increased F-measure scores on the AS dataset from 67.9 to 78.4 and the CITYU dataset from 69.1 to 82.5.

As shown in Table \ref{tab:q&g}, Qwen1.5-7B-Chat surpassed GPT models in tasks related to Chinese language processing, yet demonstrated less effectiveness in segmenting other languages. Qwen series occasionally produces translations into Simplified Chinese for words from other languages. Interestingly, the conversion from Traditional to Simplified Chinese can also be regarded as a form of "translation". This behavior does not imply a lack of comprehension by the Qwen1.5 models. Rather, it suggests that they can process how these "foreign" languages are structured and segmented. This kind of hallucination typically occurs in models where the primary pre-training corpora do not align with the target language they are tasked to segment. This observation underscores how the pre-training texts significantly influence LLMs' interpretation of prompts, akin to how a non-native English speaker might initially think in their mother tongue before translating thoughts into English. It highlights a shift from the traditional sequence of "segment first, comprehend later" to a more integrated approach where LLMs "comprehend first, segment later". This shift indicates a potential advancement in how language models process and understand text, integrating comprehension and segmentation in a more dynamic manner.

It's important to note that the Qwen1.5-32B-Chat model was released after the other models in the Qwen1.5 series as an intermediate option between the 14B and 72B models. Its architecture and training methodologies exhibit some variations from the other models in the series. These differences might explain why the Qwen1.5-32B-Chat model shows a slight decrease in performance on certain datasets compared to the 7B and 14B models. With the consistent architecture and an increase in parameters, F-measure generally improves across different languages, demonstrating that larger LLMs possess enhanced capabilities for generalization and comprehension in NLP. This trend indicates that the expansion in model capacity not only aids in handling more complex linguistic features but also improves adaptability to varied linguistic contexts. 

\subsection{From LLM-WS to LLACA}

Although LLMs demonstrate capabilities in word segmentation, employing them for this specific NLP task is impractical due to the significant time and economic costs involved. From the results shown in Table \ref{tab:res} and \ref{tab:time}, it is evident that out proposed approach LLACA not only offers a significantly faster inference speed but also maintains the statistical integrity of the patterns recognized by the LLM, often surpassing the LLM’s performance. This improvement is largely attributed to LLACA's ability to eliminate hallucinations generated by the LLM. 

\input{tab/time.tex}

Figure \ref{fig:comp} illustrates the evolution of four lexicon-based methods inspired by Qwen1.5-110B-Chat over several iterations. \textbf{LLACA}: Our proposed 
 method with variable \(n\)-gram possibility model, incorporating Viterbi decoding.
\textbf{LLM-WS-Uni}: A straightforward unigram model, also incorporating Viterbi decoding. \textbf{LLM-WS-FMM}: The Forward Maximum Matching algorithm. \textbf{LLM-WS-BMM}: The Backward Maximum Matching algorithm.

As the iterations progress, the four methods begin to plateau. The two statistical methods (LLACA, LLM-WS-Uni) consistently outperform the two greedy methods (LLM-WS-FMM, LLM-WS-BMM), as they account more thoroughly for global states rather than merely local matches. This outcome supports the premise that LLACA's enhancements to LLM outputs are not merely due to an expanded lexicon, but also because of the effective utilization of latent semantic information represented in statistical form. Among the statistical methods, our LLACA integreted with variable \(n\)-gram possibility model, as detailed in Equation \ref{eqn:trans_prob}, consistently excels over the simpler unigram model. This suggests its rationality in modeling the probabilities of natural language since ambiguity often occurs between similar patterns.

Besides, we observed that the performances with  Qwen1.5-110B-Chat was suboptimal on the Japanese dataset KWDLC. The presence of noise within the data patterns contributed to a slight degradation in performance as the model iterations progressed. Our LLACA was the least affected by the noisy data patterns and maintained a relatively good F-measure score overall. This suggests that LLACA is more robust to inputs with noises.

\subsection{On OOV Handling}

The tokenization of LLMs is based on UTF-8 code but not supervised vocabulary, and the LLM segments the words based on its comprehendion, there's no OOV issues of LLMs ideally. Our LLM-based approach LLACA actually offers advantages in handling OOV words: \textbf{1)} The dictionary can be dynamically supplemented for different domains through our unsupervised approach. \textbf{2)} This allows better domain adaptation and OOV handling compared to static dictionary approaches.

Considering that the most frequently used words are shared in multiple scenes, our method could handle the common part without forgetting them. Here we represent the comparison between SLM~\cite{sun@SLM} and LLACA (inspired by Qwen1.5-7B) in Table~\ref{tab:oov}.

\input{tab/oov.tex}

Note that the SLM results listed here are from our re-train results based on its official implementation. To make a fair comparison, the unsupervised training process used only unsegmented test corpus the same as ours, thus the results might be different from what they are represented in the original paper of SLM. With ideally the same vocabulary, our approach shows consistent better performance across different domains.

To further substantiate the validity of the variable $n$-gram approach employed by our LLACA model, we have also compared with various modeling techniques. Detailed discussions on these comparisons can be found in the Appendix \ref{ac}.

\section{Conclusion}
\label{conclu}

In this paper, we explore the word segmentation capabilities of LLMs. We conclude that the task of word segmentation can serve as an effective measure of an LLM's ability to comprehend prompts and apply logical reasoning in natural languages. To fully harness the deep comprehension capabilities of LLMs, our proposed method LLACA integrates the rapid pattern recognition of the AC automaton with a novel variable $n$-gram model, surpassing previous benchmarks and setting new state-of-the-art unsupervised results. 

\section*{Broader Impacts}
\label{impact}
Word segmentation is not a standalone task. The ambiguities need to be resolved through the context. Previous segmental models lacked the overall comprehension of the entire text, let alone multimodal comprehension. In the era of LLMs, LLM-WS takes a radically different approach, which can leverage long-range context and even multimodal associations. This emergent ability has transformed word segmentation from a pure statistical machine learning problem to a new paradigm. Building on the capabilities of LLMs, we propose LLACA, which can infer faster and perform better than previous unsupervised word segmentation methods. We hope our work will inspire more LLM-based NLP researches and applications. 

\section*{Limitations}
\label{lim}
Now that word segmentation has evolved into a comprehension task, the upper limit may lie in the inherent inconsistencies of word segmentation itself. Some segmentation results from LLMs may accurately reflect the language understanding, yet they may not align with the traditional "golden standard" annotations. To better assess the language comprehension capabilities of LLMs in NLP tasks, new evaluation standards should be developed to align with the current paradigm shift. The existing standards, which were designed for previous segmentation models, may no longer adequately capture the nuanced understanding exhibited by LLM-based approaches.

\bibliography{custom}

\appendix

\input{appendix.tex}

\end{document}

%% file: tab/main.tex
\begin{table*}[htbp]
    \centering
    \setlength{\tabcolsep}{3.3pt}
    \footnotesize
    \caption{F-measure on multilingual
    datasets compared with previous state-of-the-art models. "Uni" refers to the integration of LLM-WS and the unigram probability model, while "LLACA" denotes our newly proposed method. The asterisk (*) indicates evaluations performed after conversion from Traditional to Simplified Chinese using OpenCC. And the dagger ($\dag$) means the results are not directly comparable due to variations in the processing of the dataset. LLMs listed here are all Chat version.}
    \label{tab:res}
    \begin{tabular}{@{} llccccccccc @{}}
    \toprule
     \multicolumn{2}{c}{\multirow{2}*{\textbf{Method}}} & \multicolumn{4}{c}{Chinese} & \multicolumn{2}{c}{Japanese} & Korean & \multicolumn{2}{c}{Thai}\\
    \cmidrule(l{6pt}r{6pt}){3-6} \cmidrule(l{6pt}r{6pt}){7-8} \cmidrule(l{6pt}r{6pt}){9-9} \cmidrule(l{6pt}r{0pt}){10-11}
     & & \textbf{AS} & \textbf{CITYU} & \textbf{MSR} & \textbf{PKU} & \textbf{KWDLC} & \textbf{UD\_JA} & \textbf{UD\_KO} & \textbf{BEST} & \textbf{UD\_TH}\\
    \midrule
    \multirow{10}*{Baselines} & nVBE~\citeyearpar{magistry@nVBE} & 76.6 & 76.7 & 81.3 & 80.0 & - & - & - & - & --\\
    & NPY-2~\citeyearpar{mochihashi@NPY} & - & 82.4 & 80.2 & - & - & - & - & - & - \\
    & NPY-3 & - & 81.7 & 80.7 & - & - & - & - & - & - \\
    & Joint~\citeyearpar{chen@joint} & - & - & 81.7 & 81.1 & - & - & - & - & - \\
    & SLM-2~\citeyearpar{sun@SLM} & 79.4 & 78.2 & 78.5 & 80.2 & - & - & - & - & - \\
    & SLM-3 & 80.3 & 80.5 & 79.4 & 79.8 & - & - & - & - & - \\
    & SGB-A-12~\citeyearpar{wang2022unsupervised} & - & - & - & - & - & -  & - & $80.1^\dag$ & -\\
    & SGB-C-4 & 81.0 & 80.0 & 74.0 & 80.0 & - & - & - & - & - \\
    & SGB-C-5 & \textbf{82.4} & 78.5 & 80.4 & 78.4 & - & - & - & - & - \\
    & PYHSMM~\citeyearpar{PYHSMM} & - & \textbf{82.6} & \textbf{82.9} & \textbf{81.6} & - & - & - & $\textbf{82.1}^\dag$ & -\\
    \midrule
    \multirow{3}*{Qwen1.5-7B} & LLM & 78.4* & 82.5* & 84.2 & 86.7 & 52.7 & 49.8 & 36.8 & 8.9 & 21.3\\
    & Uni & 84.6* & 85.9* & 86.4 & 87.5 & 64.9 & 58.8 & 43.2 & 37.5 & 32.7\\
    & LLACA & \textbf{84.8}* & \textbf{86.5}* & \textbf{86.7} & \textbf{87.7} & \textbf{66.4} & \textbf{62.8} & \textbf{48.3} & \textbf{48.6} & \textbf{39.2}\\
    \midrule
    \multirow{2}*{Qwen1.5-14B} & LLM & 78.0 & 72.1 & 85.7 & 87.0 & 70.3 & 64.6 & 43.9 & 48.7 & 57.1\\
    & LLACA & \textbf{86.6} & \textbf{82.7} & \textbf{87.8} & \textbf{89.3} & \textbf{76.7} & \textbf{69.2} & \textbf{51.9} & \textbf{64.7} & \textbf{62.2}\\
    \midrule
    \multirow{2}*{Qwen1.5-32B} & LLM & 80.5 & 74.9 & 85.9 & 86.3 & 67.2 & 64.3 & 48.3 & 45.9 & 53.8\\
    & LLACA & \textbf{86.8} & \textbf{84.0} & \textbf{87.6} & \textbf{87.6} & \textbf{74.1} & \textbf{68.0} & \textbf{54.9} & \textbf{64.9} & \textbf{61.4}\\
    \midrule
    Qwen1.5-72B & LLACA & \textbf{88.3} & \textbf{88.1} & \textbf{88.2} & \textbf{88.7} & \textbf{76.1} & \textbf{69.4} & \textbf{58.9} & \textbf{68.9} & \textbf{70.0}\\
    \bottomrule
    \end{tabular}
\end{table*}

%% file: tab/qwen_gpt.tex
\begin{table*}[htbp]
    \centering
    \setlength{\tabcolsep}{3.3pt}
    \footnotesize
    \caption{The comparison between Qwen1.5-7B-Chat and GPT-4.}
    \label{tab:q&g}
    \begin{tabular}{@{} lccccccccc @{}}
    \toprule
    \textbf{Model} & \textbf{AS} & \textbf{CITYU} & \textbf{MSR} & \textbf{PKU} & \textbf{KWDLC} & \textbf{UD\_JA} & \textbf{UD\_KO} & \textbf{BEST} & \textbf{UD\_TH}\\
    \midrule
    GPT-4 & 76.5 & 80.6 & 75.0 & 76.3 & \textbf{79.7} & \textbf{78.7} & \textbf{44.9} & \textbf{76.3} & \textbf{75.1}\\
    Qwen1.5-7B-Chat & \textbf{78.4}* & \textbf{82.5}* & \textbf{84.2} & \textbf{86.7} & 52.7 & 49.8 & 36.8 & 8.9 & 21.3\\
    \bottomrule
    \end{tabular}
\end{table*}

%% file: tab/time.tex
\begin{table}[htbp]
    \centering
    \setlength{\tabcolsep}{3.3pt}
    \footnotesize
    \caption{The comparison of the average inference time and F1 scores between Qwen1.5-14B-Chat and LLACA inspired by it over 4 iterations.}
    \label{tab:time}
    \begin{tabular}{@{} lcc @{}}
    \toprule
    \textbf{Model} & \textbf{Time$\downarrow$ }(MSR) & \textbf{$F\uparrow$ }(MSR) \\
    \midrule
    Qwen1.5-14B-Chat & 3.65h & 85.9\\
    LLACA & 2.01s & 87.7\\
    \bottomrule
    \toprule
    \textbf{Model} & \textbf{Time$\downarrow$ }(PKU) & \textbf{$F\uparrow$ }(PKU) \\
    \midrule
    Qwen1.5-14B-Chat & 2.85h & 87.5\\
    LLACA & 1.80s & 88.9\\
    \bottomrule
    \end{tabular}
\end{table}

%% file: tab/oov.tex
\begin{table}[htbp]
    \centering
    \caption{Comparison between SLM~\cite{sun@SLM} and LLACA on OOV Handling for MSR, PKU, and CTB Datasets}
    \label{tab:oov}
    \begin{tabular}{@{} lccc @{}}
    \toprule
    \multirow{2}*{Training-Model} & \multicolumn{3}{c}{Test}\\
    \cmidrule(l{6pt}r{6pt}){2-4}
     & \textbf{MSR} & \textbf{PKU} & \textbf{CTB} \\
    \midrule
    MSR-SLM-3 & 73.9 & 69.8 & 67.4 \\
    MSR-LLACA & \textbf{86.7} & \textbf{75.2} & \textbf{70.0} \\
    PKU-SLM-3 & 70.8 & 76.6 & 69.2 \\
    PKU-LLACA & \textbf{78.2} & \textbf{87.7} & \textbf{72.9} \\
    CTB-SLM-3 & 69.7 & 70.1 & 76.0 \\
    CTB-LLACA & \textbf{77.0} & \textbf{75.5} & \textbf{88.0} \\
    \bottomrule
    \end{tabular}
\end{table}

%% file: appendix.tex
\section{Experimental Details}
\label{exp_dtl}
\subsection{Algorithm}
It is feasible to use the LLACA automaton independently of the LLM after its initial construction. This flexibility allows for manual additions to the vocabulary or for continued refinement and "distillation" from existing LLM outputs. As LLMs are exceptionally effective at exploring and identifying new vocabulary—acting as advanced "word explorers"—the issue of handling out-of-distribution data is significantly mitigated. We can always tailor the LLACA to be domain-specific, thus maintaining its effectiveness across varying data distributions. Algorithm \ref{alg:cut} describes the overall procedure including vocabulary construction and word segmentation.

Furthermore, the design of LLACA is inherently extensible, with each recognized pattern functioning as a weighted edge within the automaton. This structure facilitates the integration of additional patterns and rules, such as regular expressions for identifying URLs or other specialized formats. By interpolating the weights assigned to these new patterns, LLACA can be adapted to recognize and process a broader array of textual features, enhancing its utility and applicability.

\input{tab/datasets.tex}

\subsection{Datasets}
\paragraph{Chinese} We utilized four standard datasets from the SIGHAN Bakeoff 2005~\cite{emerson2005second}, widely recognized in computational linguistics research: AS, CITYU, MSR, and PKU. The AS and CITYU datasets contain Traditional Chinese texts, while MSR and PKU are composed of Simplified Chinese texts. These datasets provide a robust foundation for evaluating segmentation performance across various forms of Chinese script.
\paragraph{Japanese} For Japanese, we employed the Kyoto University Word Dependency Corpus (KWDLC)~\cite{hangyo-etal-2012-building} and the Universal Dependencies (UD) Japanese GSD treebank~\cite{nivre2020universal}, which are instrumental in studying segmentation in scripts without clear delimiters.
\paragraph{Korean} For Korean, we employed the Universal Dependencies (UD) Korean GSD treebank~\cite{nivre2020universal}. Additionally, we deleted the spaces in the original test dataset.
\paragraph{Thai} For Thai, we utilized the BEST dataset~\cite{kosawat2009best} and the UD Thai PUD treebank~\cite{nivre2020universal}. Additionally, we created a subset by randomly selecting 1,000 sentences from the BEST-Novel dataset's training data. All delimiters were removed from this subset to prepare it for use as both training and testing data.

\paragraph{Statistics and Licenses for Datasets} Table \ref{tab:datasets} provides a summary of the datasets used for evaluation, while Table \ref{tab:licenses} details the licenses associated with each dataset.

Except for BEST was a subuset contained 1,000 sentences randomly sampled with random seed 17 by \verb|numpy.random.choice| from the BEST-Novel training dataset, all other test datasets maintained their original splits. We only used their test datasets for LLM-WS and testing.

\input{tab/licenses.tex}

\subsection{Experimental Environment}
For close-sourced models GPT-3.5-Turbo and GPT-4, we employed them by API. And for other open-sourced models, we initialized them from pre-trained checkpoints and employed them on 8 A100-SXM80GB GPUs.

\subsection{Parameters}
We set the top ratio to a conservative level of 0.99, allowing LLACA to incorporate a broader range of words into the analysis. 


\section{Baselines}
\label{base}
The detail of baseline models is following:
\begin{itemize}
    \item $n$VBE~\cite{magistry@nVBE}: a system utilizing Normalized Variation of Branching Entropy.
    \item NPY-$n$~\cite{mochihashi@NPY}: a nested $n$-gram hierarchical Pitman-Yor language model, where Pitman-Yor spelling model is embedded in the word model.
    \item Joint~\cite{chen@joint}: the “HDP+HMM” model initialized with $n$VBE model.
    \item SLM-$n$~\cite{sun@SLM}: the first neural model for unsupervised CWS, where $n$ denotes the maximum word length. 
    \item SGB-A, SGB-C~\cite{wang2022unsupervised}: a model maximizing the generation probability of a sentence given its all possible segmentation, where A and C denote 2 different decoding algorithms.
    \item PYHSMM~\cite{PYHSMM}: a nonparametric Bayesian model for joint unsupervised word segmentation and part-of-speech tagging from raw strings.
\end{itemize}

\begin{algorithm*}
\setlength{\tabcolsep}{3.3pt}
\footnotesize
\caption{Word Segmentation with LLACA}
\label{alg:cut}
\begin{algorithmic}[1] 
\State \textbf{Input:} Raw text data
\State \textbf{Output:} Segmented text
\Procedure{LLACA}{text}
    \State Randomly shuffle the raw text into batches
    \For{each batch}
        \State Get patterns from LLM-WS
        \State Apply PMI filtering to refine patterns
        \State Add filtered patterns to LLACA
    \EndFor
    \State Pre-process with normal patterns like number, alpha and symbols
    \State Initialize Viterbi probabilities based on patterns in LLACA using Equation \ref{eqn:trans_prob}
    \State Prepare a table to track paths and their probabilities
    \For{each character index \( i \) in text}
        \For{each state \( s \) representing a possible pattern ending at \( i \)}
            \State Calculate the highest probability path ending in \( s \) using:
            \State \( \max_{s'} \log P(s' \rightarrow s) + \log P(s') \)
            \State where \( s' \) is a state leading to \( s \) and \( P \) are the transition probabilities
            \State Store this path if it has the highest log probability for \( s \)
        \EndFor
    \EndFor
    \State Backtrack from last character to find the path with the highest probability
    \State \textbf{return} Segmented text based on the best path
\EndProcedure
\end{algorithmic}
\end{algorithm*}

\section{Analysis of the Limits of Unsupervised Word Segmentation}

As indicated by \cite{huang@CWSreview, zhao2008empirical}, standard inconsistencies occur across different datasets for word segmentation. Therefore, it is essential to consider consistency across these datasets as the upper limit for evaluating unsupervised segmentation methods. We selected three Simplified Chinese datasets and pre-trained segmental models on each to replay the experiment in \cite{huang@CWSreview}. The model architecture is adapted from \cite{wang-xu-2017-convolutional} and we utilizes pre-trained models provided by HanLP\footnote{\url{https://github.com/hankcs/HanLP/tree/master}}.

Let the F-measure of each model on its respective test dataset be denoted as $F_0$. Typically, supervised models perform optimally on their corresponding test sets but exhibit diminished performance on others. Hence, we use $F_0$ as a normalization factor to measure consistency across datasets. The values of $F_0$ are 95.4 (CTB), 97.1 (MSR), and 95.5 (PKU). Each F-measure was normalized by the $F_0$ of its training dataset, achieving a consistency score of 1.0 for each model on its test dataset.

\input{tab/consistency.tex}

As shown in Table \ref{tab:consis}, the lowest consistency rate observed is 86.5\%, while the highest is 94.7\%, with an average consistency of 93.8\%. Some previous studies have regarded that the lowest consistency rate is the ceiling for unsupervised word segmentation methods~\cite{sun@SLM, PYHSMM}. We propose that it is more reasonable to estimate the ceiling as the average consistency. Utilizing Qwen1.5-110B-Chat, our LLACA achieved a 90.3 F-measure on the CITYU dataset.

Aside from the inconsistency of segmentation criteria, most traditional unsupervised word segmentation methods are based on boosting. However, our approach is based on the semantic understanding capabilities of LLMs. Therefore, our method has re-explored the upper limit of unsupervised word segmentation, which also demonstrates that our approach has ushered in a new era for the development of unsupervised word segmentation.

\begin{figure*}[htbp]
    \centering
    \includegraphics[width=1\linewidth]{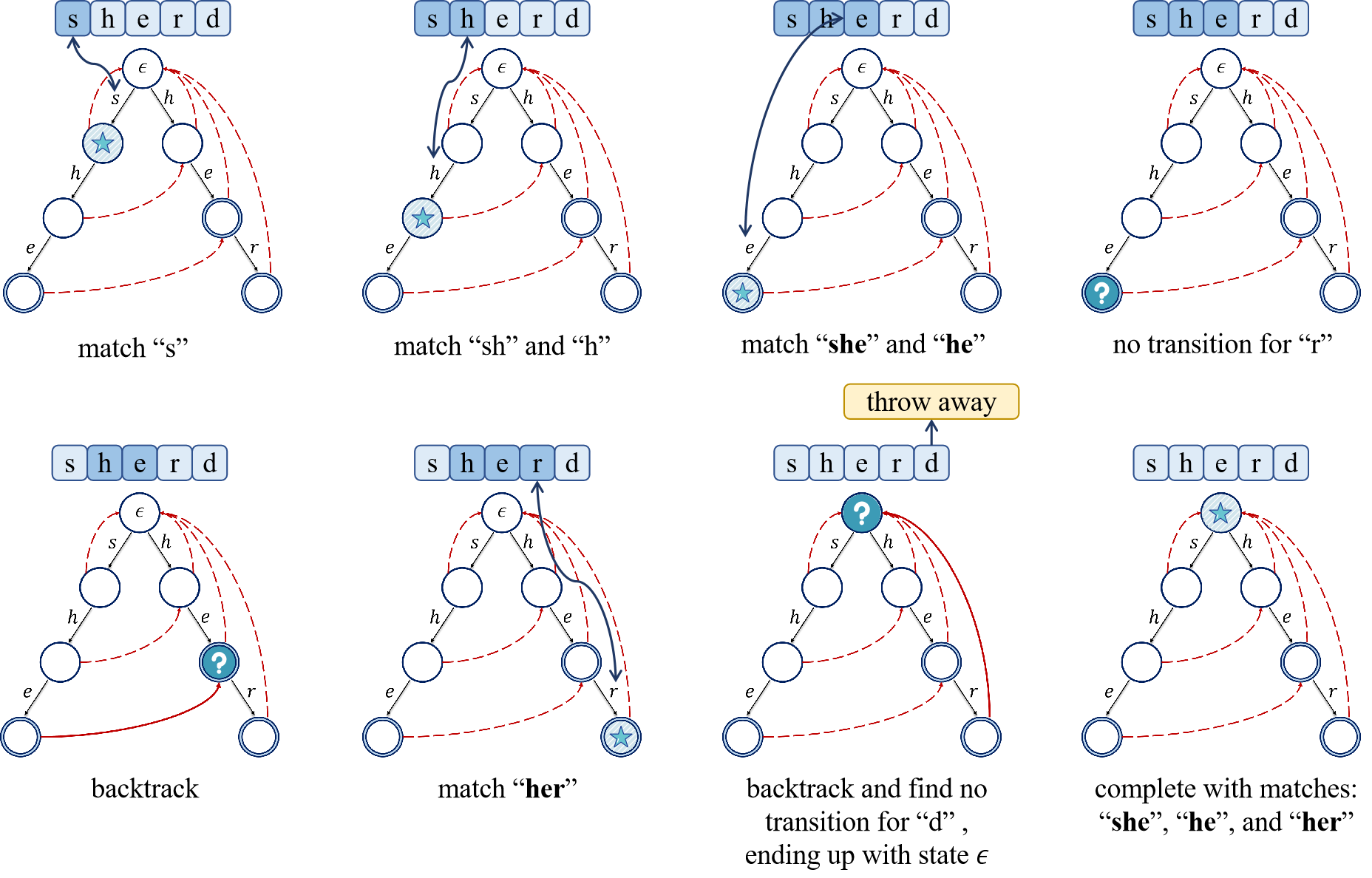}
    \caption{The process of how "sherd" matches patterns "she", "he", and "her" by transitions on an AC automaton.}
    \label{fig:ac-trans}
\end{figure*}

\section{Aho-Corasick Automaton} 
Once the Trie is assembled, failure links are established. Each failure link at a node connects to the longest proper suffix of the string at that node, which also serves as a prefix for another pattern in the Trie. If no such suffix exists, the link reverts to the root. This is analogous to the "failure function" in the Knuth-Morris-Pratt (KMP) string-matching algorithm~\cite{knuth1977fast}, but Aho-Corasick extends this idea to work efficiently for multiple patterns. 

Figure \ref{fig:automaton} illustrates the AC automaton’s structure, showcasing failure links in red and final states with double circles, though some transitions may be omitted for clarity. The process begins with an input sequence that progresses through the Trie. If a mismatch occurs, such as when at the state "she" and the next input is "r" without a corresponding edge, the automaton utilizes failure links to backtrack until a valid node with the "r" edge is found or until it returns to the root. When the automaton reaches the state "her", not only is the pattern "her" itself recognized but the state of its failure pointer is also included. This forms part of a recursive process: matching a state involves sequentially matching the state of its failure pointer until it traces back to the root node, which represents the absence of further matches, denoted as $\epsilon$.

For further clarification, Figure~\ref{fig:ac-trans} illustrates the matching process for the string "sherd", identifying the substrings "she", "he", and "her". The elements highlighted in dark blue represent both the longest pattern prefix that the current state can match, and the minimal suffix information necessary for subsequent matches. This setup can be visualized as a "sliding window" that moves from left to right across each character. During normal transitions, this sliding window accordingly steps to the right. Conversely, during backtracking, the state transitions via the failure pointer, effectively discarding any irrelevant left-side components. Note that in actual implementations, the failure links in AC automata are primarily used during the construction phase and the match phase. Once the automaton is constructed, these failure links are often replaced by virtual transitions that directly lead to the correct states like Algorithm~\ref{alg:trans}. This optimization streamlines the matching process, enhancing efficiency by reducing unnecessary transitions.

The double-array Trie~\cite{aoe1989efficient}, markedly enhances the efficiency of space utilization in Trie data structures. This innovation reduces the memory footprint traditionally associated with Trie implementations. As a result, the double-array-based Aho-Corasick automaton emerges as an almost ideal solution for applications involving multiple pattern matching due to its optimal balance of space efficiency and performance.

\input{tab/ord}

\section{Further Validations for LLACA}
\label{ac}

\subsection{Probability Modeling}
LLACA employs a variable $n$-gram approach to model probabilities at a global level. This method not only captures local dependencies but also considers global contextual information. Let $p_i$ represent the probability that the model assigns to the segmentation of the $i$-th sentence $s_i$. The perplexity of one sentence can be defined as follows, where $n_i$ denotes the length of $s_i$:
\begin{subequations}
\begin{align}
    \PPL(s_i) &=  \sqrt[n_i]{\frac{1}{p_i}} \\
    & = \exp\left(-\frac{\log p_i}{n_i}\right)
\end{align}
\end{subequations}

For the total texts, we still adopt geometric mean. The perplexity for $\mathbf{t}=[s_1, s_2, \cdots, s_k]$ could be measured as below:
\begin{subequations}
\begin{align}
    \PPL(\mathbf{t}) & = \sqrt[k]{\PPL(s_1)\PPL(s_2)\cdots\PPL(s_k)}\\
    & = \sqrt[k]{\Pi_{i=1}^k\exp\left(-\frac{\log p_i}{n_i}\right)}\\
    & = \exp\left(\frac{\sum_{i=1}^n-\frac{\log p_i}{n_i}}{k}\right)
\end{align}
\end{subequations}

\subsection{Baselines}

Tools mentioned in the discussion are available in GitHub. Details of them are below:

\begin{itemize}
    \item Jieba\footnote{\url{https://github.com/fxsjy/jieba}}: Jieba is a highly popular Chinese text segmentation tool known for its ease of use and versatility. Jieba also allows for custom dictionary integration, making it adaptable for specific vocabularies or industry terms.
    \item MeCab\footnote{\url{https://github.com/taku910/mecab}}: MeCab is a sophisticated morphological analyzer for Japanese text, implemented in C++ and employing Conditional Random Fields (CRF) as its core algorithm.
    \item Komoran (in KoNLPy\footnote{\url{https://github.com/konlpy/konlpy}}): As part of the KoNLPy suite, Komoran is tailored for Korean text segmentation. It is particularly noted for its accuracy in analyzing formal and well-structured documents. Komoran is suitable for academic and professional applications where precision is crucial.
    \item AttaCut (in PyThaiNLP\footnote{\url{https://github.com/PyThaiNLP/pythainlp}}): A modern tool designed specifically for Thai, AttaCut is embedded in PyThaiNLP project. It uses deep learning models, particularly CNNs, to segment Thai text, which is known for its absence of clear word boundaries.
    \item NewMM (in PyThaiNLP): Another engine embedded in PyThaiNLP, NewMM is a multi-dictionary-based maximizer matching algorithm that efficiently handles WS in Thai. It's designed to be fast and is the default engine in PyThaiNLP due to its robustness in general-purpose applications.
\end{itemize}

\subsection{Starting from an Ordinary Dictionary}

Starting with an "ordinary dictionary" trained on the \textit{People's Daily} corpus provided by
Jieba, further expanded by additional dictionaries extracted from raw texts in the MSR dataset using
Qwen1.5-7B to Qwen1.5-72B. The results, evaluated using F1 scores on the MSR dataset, are presented in Table \ref{tab:ord}.

\input{tab/cmp}

\subsection{Comparisons with "Off-the-Shelf" Tools}

Table \ref{tab:cmp} presents comparisons with other off-the-shelf tools in Japanese, Korean and Thai. While these results are not directly comparable, they serve as an indicator of the limitations inherent in supervised methods and the advantages of our unsupervised approach. Supervised methods often struggle with out-of-domain (OOD) and out-of-vocabulary (OOV) issues. However, in practical scenarios, labeled domain-specific data is not always abundantly available. Utilizing LLMs as our "word explorer", similar to human cognitive processes, ideally eliminates concerns related to OOD and OOV. Many supervised methods have not transitioned into practical tools due to concerns over efficiency and the ability for customization. Here, we demonstrate the transferability and high efficiency of LLACA, underscoring its capability to be effectively implemented in real-world applications. We are confident in and committed to the potential of LLACA.

\input{tab/perplexity.tex}

Table \ref{tab:ppl} shows the perplexity comparison between LLM-WS-Uni and LLACA. On four datasets, LLACA achieved higher F-measure and lower perplexity compared to LLM-WS-Uni. Higher F-measure and lower perplexity represent a higher probability of the word segmentation results, implying that the word segmentation results are more reasonable.

\begin{algorithm*}
\caption{Calculate fail links and virtual transitions for nodes in an AC automaton}
\label{alg:trans}
\begin{algorithmic}[1]
\Function{get\_transitions}{}
    \State Initialize an empty queue $Q$
    \For{$v \gets$ children of the root ($\epsilon$)}
        \State fail($v$) $\gets \epsilon$ \Comment{Set initial fail state to $\epsilon$}
        \State Enqueue $v$ into $Q$
    \EndFor
    \While{not $Q$.isEmpty()}
        \State $u \gets Q$.dequeue()
        \For{$i \gets$ possible transitions from $u$}
            \State $v \gets$ child($u$, $i$)
            \If{$v \neq \epsilon$}
                \State fail($v$) $\gets$ child(fail($u$), $i$) \Comment{Update the failure pointer}
                \State Enqueue $v$ into $Q$
            \Else
                \State child($u$, $i$) $\gets$ child(fail($u$), $i$) \Comment{Set virtual transition}
            \EndIf
        \EndFor
    \EndWhile
\EndFunction
\end{algorithmic}
\end{algorithm*}

\section{Discussions}

\subsection{With LLMs, Why We Still Need Word Segmentation (WS)}

BPE tokenizers, especially the one used in OpenAI's Tiktoken (also employed by Qwen-1.5), operate at a \textit{byte-level} and do not inherently understand semantic boundaries. This means that while models like GPT and Qwen-1.5 perform impressively on many tasks, their understanding is based on statistical co-occurrence rather than semantic comprehension. The impressive performance of these models can largely be attributed to their well-designed architectures and extensive pre-training.

However, this comes at a cost. The high time and space complexity of inference with such large models can be prohibitive, particularly in environments where computational resources are limited. Furthermore, according to the "No Free Lunch Theorem", these models may still lag behind in domain-specific tasks compared to specialized NLP tools. Additionally, smaller models are often prone to hallucinations due to their reduced capacity and generalist training.

Given these considerations, the task of WS remains critically important for several reasons:

\begin{itemize}
    \item \textbf{Efficiency}: Many NLP tasks require high computational efficiency. Fast and effective WS tools can provide essential semantic information more quickly and with fewer resources than LLMs. For example, some rule-based methods could greatly benefit from this.
    \item \textbf{Accessibility}: Not all researchers and developers have the means to deploy and maintain LLMs like Qwen efficiently. By continuing to develop and improve WS techniques, we ensure that robust, less resource-intensive solutions are available, keeping the field of NLP inclusive and versatile.
    \item \textbf{Reliability}: Effective segmentation reduces errors in further processing steps, such as parsing and translation, ensuring that the output is both accurate and contextually appropriate. This is crucial in professional settings where precision is paramount, such as legal and medical document analysis.
\end{itemize}

In summary, while large pre-trained models offer broad capabilities, WS tasks play a crucial role in achieving high accuracy and efficiency in specific applications, ensuring that CJK NLP technology remains accessible and practical across a diverse range of use cases.

\subsection{Why We Still Need Unsupervised Word Segmentation}

Generally, we consider unsupervised methods from these aspects:

\begin{itemize}
    \item \textbf{Data Availability and Cost}: Acquiring labeled data is frequently costly and time-consuming. In many practical scenarios, such data may not even be available. Unsupervised learning, on the other hand, does not require labeled inputs and can be applied directly to raw data. This makes it particularly valuable in situations where data labeling is impractical or too expensive.
    \item \textbf{Pattern Discovery}: Unsupervised learning excels at discovering hidden patterns and structures in data that are not initially evident. For instance, clustering algorithms can reveal intrinsic groupings and structures within the data that supervised methods might overlook because they focus solely on the target outcomes defined by the labeled data.
    \item \textbf{Flexibility and Adaptability}: In dynamic environments where data distributions change over time, supervised models may require retraining with new labeled data, which can be both costly and impractical. Unsupervised learning models, can adapt to changes in input data without needing completely new labels.
\end{itemize}

Regarding the WS task, we introduce LLACA as a practical solution to maintain the WS capabilities of LLMs for actual use. This approach can significantly benefit various NLP tasks and supervised models by providing semantic information derived from words. Furthermore, LLACA naturally adapt to changes in domain and era, making it a versatile tool in the evolving landscape of NLP.

%% file: tab/datasets.tex
\begin{table*}[htbp]
    \centering
    \setlength{\tabcolsep}{3.3pt}
    \footnotesize
    \caption{The statistics overview of the datasets used in the evaluation across different languages: Chinese (ZH), Japanese (JA), Korean(KO), and Thai (TH).}
    \label{tab:datasets}
    \begin{tabular}{@{} cccccccc @{}}
    \toprule
    \textbf{Language} & \textbf{Dataset} & \textbf{Test Size (KB)} & \textbf{Total Tokens} & \textbf{Total Chars} & \textbf{Unique Tokens} & \textbf{Unique Chars}\\
    \midrule
    \multirow{4}*{ZH} & AS & 603 & 122610 & 197681 & 18811 & 3707\\
     & CITYU & 196 & 40936 & 67690 & 9001 & 2702\\
     & MSR & 547 & 106873 & 184355 & 12923 & 2838\\
     & PKU & 497 & 104372 & 172733 & 13148 & 2934\\
    \midrule   
    \multirow{2}*{JA} & KWDLC & 194 & 35869 & 64905 & 6144 & 1821\\
     & UD\_JA & 62 & 13034 & 21322 & 3568 & 1494\\
    \midrule
    \multirow{1}*{KO} & UD\_KO & 100 & 11677 & 32742 & 7102 & 1125\\
    \midrule
    \multirow{2}*{TH} & BEST & 328 & 31697 & 112261 & 3804 & 102\\
     & UD\_TH & 282 & 22322 & 96161 & 4047 & 134\\
    \bottomrule
    \end{tabular}
\end{table*}

%% file: tab/licenses.tex
\begin{table}[htbp]
    \centering
    \footnotesize
    \caption{Licenses of the datasets.}
    \label{tab:licenses}
    \begin{tabular}{@{} lc @{}}
    \toprule
    \textbf{Dataset} & \textbf{License}\\
    \midrule
    AS & Research Purpose\\
    CITYU & Research Purpose\\
    MSR & Research Purpose\\
    PKU & Research Purpose\\
    KWDLC & Research Purpose\\
    UD\_JA & CC BY-SA 4.0\\
    UD\_KO & CC BY-SA 4.0\\
    BEST & CC BY-NC-SA 3.0\\
    UD\_TH & CC BY-SA 3.0\\
    \bottomrule
    \end{tabular}
\end{table}

%% file: tab/consistency.tex
\begin{table}[htbp]
    \centering
    \caption{Consistency rate among Simplified Chinese datasets CTB, MSR and PKU}
    \label{tab:consis}
    \begin{tabular}{@{} lccc @{}}
    \toprule
    \multirow{2}*{Test} & \multicolumn{3}{c}{Trainig}\\
    \cmidrule(l{6pt}r{6pt}){2-4}
     & \textbf{CTB} & \textbf{MSR} & \textbf{PKU} \\
    \midrule
    \textbf{CTB}  & 1.000  & 0.865  & 0.944 \\
    \textbf{MSR}  & 0.871  & 1.000  & 0.947 \\
    \textbf{PKU}  & 0.933  & 0.880  & 1.000 \\
    \bottomrule
    \end{tabular}
\end{table}

%% file: tab/ord.tex
\begin{table*}[htbp]
    \centering
    \footnotesize
    \caption{Starting from an “ordinary dictionary” trained on the \textit{People's Daily} corpus provided by Jieba, further expanded by additional dictionaries extracted from raw texts in the MSR dataset using Qwen1.5-7B to Qwen1.5-72B.}
    \label{tab:ord}
    \begin{tabular}{lccccc}
        \toprule
         & Ord. & +Qwen1.5-7B & +Qwen1.5-14B & +Qwen1.5-32B & +Qwen1.5-72B\\
        \midrule
        Ours & \textbf{85.5} & \textbf{86.5} & \textbf{86.8} & \textbf{86.8} & \textbf{87.0} \\
        Jieba & 82.7 & 80.3 & 80.7 & 80.4 & 80.8\\
        \bottomrule
    \end{tabular}
\end{table*}

%% file: tab/cmp.tex
\begin{table*}[htbp]
    \centering
    \footnotesize
    \setlength{\tabcolsep}{3.3pt}
    \caption{Compared with "off-the-shelf" tools for Japanese, Korean, and Thai languages, $\dagger$ indicates that the training set may overlap with the test set used here. * denotes that we added the same training vocabulary as ours to ensure a more equitable comparison.}
    \label{tab:cmp}
    \begin{tabular}{@{} lcccccccccc @{}}
        \toprule
         & \multicolumn{2}{c}{MeCab} & \multicolumn{2}{c}{Komoran} & \multicolumn{2}{c}{AttaCut} & \multicolumn{2}{c}{NewMM} & \multicolumn{2}{c}{Ours}\\
         \cmidrule(l{6pt}r{6pt}){2-3} \cmidrule(l{6pt}r{6pt}){4-5} \cmidrule(l{6pt}r{6pt}){6-7} \cmidrule(l{6pt}r{6pt}){8-9} \cmidrule(l{6pt}r{6pt}){10-11}
         & F1 & Time (s) & F1 & Time (s) & F1 & Time (s) & F1 & Time (s) & F1 & Time (s)\\
        \midrule
        KWDLC & 88.8 & 0.06 & & & & & & & 91.9 & 0.67\\
        UD\_KO & & & 27.9$^*$ & 0.58 & & & & & 51.2 & 0.38\\
        BEST & & & & & 97.0$^\dagger$ & 14.09 & 75.3$^*$ & 0.31 & 93.5 & 1.09\\
        \bottomrule
    \end{tabular}
\end{table*}

%% file: tab/perplexity.tex
\begin{table}[htbp]
    \centering
    \setlength{\tabcolsep}{8.3pt}
    \footnotesize
    \caption{Perplexity (the lower the better) and F-measure (the larger the better) on datasets of different languages. LLM-WS-Uni and LLACA's construction were both conducted on GPT-4. }
    \label{tab:ppl}
    \begin{tabular}{@{} lcccc @{}}
    \toprule
     & \multicolumn{2}{c}{CITYU} & \multicolumn{2}{c}{MSR} \\
    \cmidrule(l{6pt}r{6pt}){2-3} \cmidrule(l{6pt}r{6pt}){4-5} 
    \textbf{Model} & F$\uparrow$ & ppl.$\downarrow$ & F$\uparrow$ & ppl.$\downarrow$ \\
    \midrule
    LLM-WS-Uni & 83.7 & 121 & 84.1 & 94 \\
    LLACA & \textbf{84.1} & \textbf{42} & \textbf{84.2} & \textbf{28}\\
    \bottomrule
    \toprule
     & \multicolumn{2}{c}{KWDLC} & \multicolumn{2}{c}{BEST} \\
    \cmidrule(l{6pt}r{6pt}){2-3} \cmidrule(l{6pt}r{6pt}){4-5} 
    \textbf{Model} & F$\uparrow$ & ppl.$\downarrow$ & F$\uparrow$ & ppl.$\downarrow$ \\
    \midrule
    LLM-WS-Uni & 82.7 & 51 & 72.1 & 14\\
    LLACA & \textbf{83.4} & \textbf{21} & \textbf{72.8} & \textbf{8}\\
    \bottomrule
    \end{tabular}
\end{table}